\documentclass[10pt,twocolumn,letterpaper]{article}

\usepackage{cvpr}              

\usepackage{graphicx}
\usepackage{amsmath}
\usepackage{amssymb}
\usepackage{booktabs}
\usepackage{times}
\usepackage{epsfig}
\usepackage{comment}
\usepackage{tabularx}
\usepackage{caption}
\usepackage{algorithm,algcompatible}
\usepackage{amssymb}
\algnewcommand\INPUT{\item[\textbf{Input:}]}
\algnewcommand\OUTPUT{\item[\textbf{Output:}]}
\usepackage{tabularray}
\usepackage{pbox}
\usepackage{adjustbox}
\usepackage{colortbl}

\usepackage[pagebackref,breaklinks,colorlinks]{hyperref}
\usepackage[capitalize]{cleveref}
\crefname{section}{Sec.}{Secs.}
\Crefname{section}{Section}{Sections}
\Crefname{table}{Table}{Tables}
\crefname{table}{Tab.}{Tabs.}
\definecolor{Gray}{gray}{0.85}
\newcolumntype{a}{>{\columncolor{Gray}}c}
\newcolumntype{b}{>{\columncolor{white}}c}

\usepackage[pagebackref,breaklinks,colorlinks]{hyperref}

\usepackage[capitalize]{cleveref}
\crefname{section}{Sec.}{Secs.}
\Crefname{section}{Section}{Sections}
\Crefname{table}{Table}{Tables}
\crefname{table}{Tab.}{Tabs.}

\def\cvprPaperID{*****} 
\def\confName{CVPR}
\def\confYear{2022}

\begin{document}

\title{Generating Reliable Pixel-Level Labels for Source Free Domain Adaptation}
\author{Gabriel Tjio\\
Centre for Frontier \\ AI Research (CFAR)\\
{\tt\small gabriel-tjio@cfar.a-star.edu.sg}
\and
Ping Liu*\\
Centre for Frontier \\ AI Research (CFAR)\\
{\tt\small liu\_ping@cfar.a-star.edu.sg}
\and
Yawei Luo\\
Zhejiang University\\
{\tt\small yaweiluo329@gmail.com}
\and
Chee Keong Kwoh\\
Nanyang Technological \\ University\\
{\tt\small asckkwoh@ntu.edu.sg}
\and
Joey Zhou Tianyi\\
Centre for Frontier \\ AI Research (CFAR)\\
{\tt\small joey\_zhou@cfar.a-star.edu.sg}
 }
\maketitle
\footnote{ * Ping Liu is the corresponding author. }
\begin{abstract}
This work addresses the challenging domain adaptation setting in which knowledge from the labelled source domain dataset is available only from the pretrained black-box segmentation model.
The pretrained  model's predictions for the target domain images are noisy because of the distributional differences between the source domain data and the target domain data.
Since the model's predictions serve as pseudo labels during self-training, the noise in the predictions impose an upper bound on model performance. 
Therefore, we propose a simple yet novel image translation workflow, ReGEN, to address this problem.
ReGEN comprises an image-to-image translation network and a segmentation network.
Our workflow generates target-like images using the noisy predictions from the original target domain images.
These target-like images are semantically consistent with the noisy model predictions and therefore can be used to train the segmentation network.
In addition to being semantically consistent with the predictions from the original target domain images, the generated target-like images are also stylistically similar to the target domain images.
This allows us to leverage the stylistic differences between the target-like images and the target domain image as an additional source of supervision while training the segmentation model. 
We evaluate our model with two benchmark domain adaptation settings and demonstrate that our approach performs favourably relative to recent state-of-the-art work. 
The source code will be made available.
\end{abstract}

\section{Introduction}
Deep learning has brought about revolutionary changes across several fields since its introduction. 
In particular, performance for computer vision tasks such as object detection\cite{Ren2015FasterRcnn}, image classification\cite{imagenet_classification} and semantic segmentation\cite{Shelhamer2014FullyCN} have all improved tremendously through the application of deep learning.
However, these advances in performance require vast amounts of labelled training data. 
While synthetic data generated with photo-realistic rendering techniques offer a potential solution for generating labelled data more easily, it has been observed that training deep learning models solely with synthetic data significantly reduces performance when tested on real-world data from the target domain.

 \begin{figure}[t]
\includegraphics[width=1.0\columnwidth,keepaspectratio]{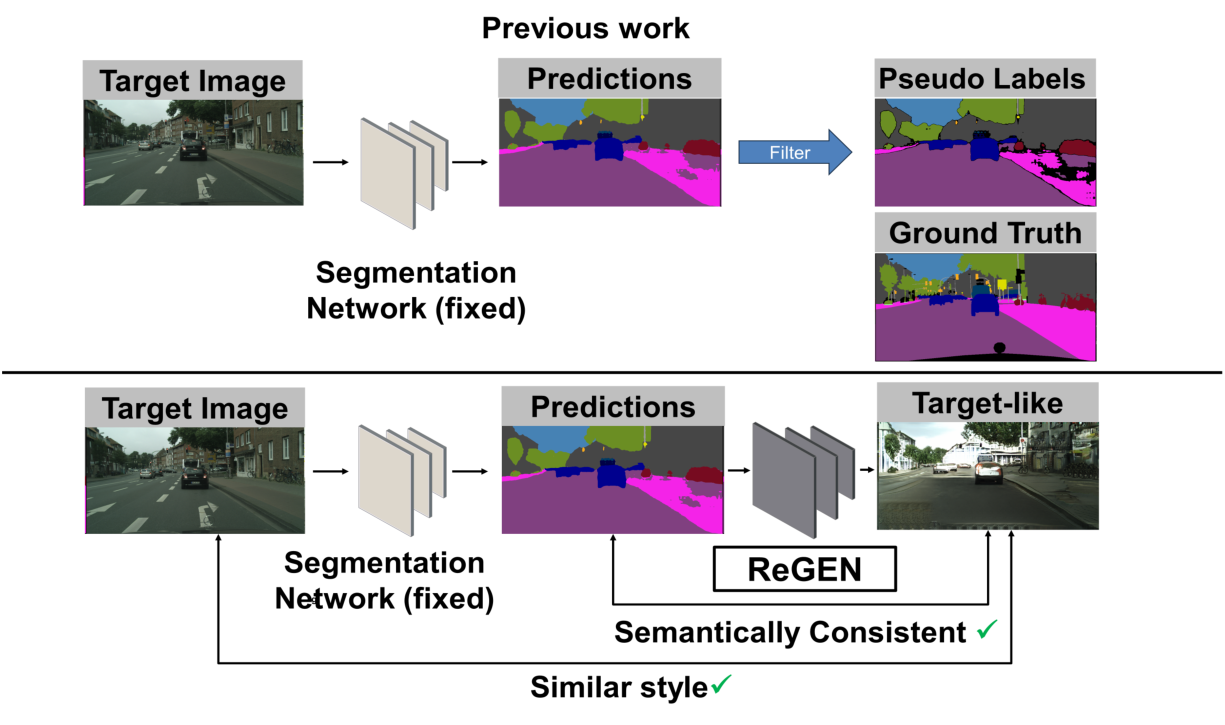}
\captionsetup{width=1.0\columnwidth}
\caption{Illustration of our approach, REGEN, compared to previous self-supervised methods. Model predictions are noisy and have to be filtered to reduce the effect of noisy labels on the model performance. Our proposed approach, ReGEN, addresses this problem by generating target-like images that are semantically consistent with the model predictions. The generated target-like images also share similar styles with the original target domain images, which minimizes the domain shift between target-like images and the target domain images.}
\label{figure:figure1_overview}
\end{figure}   


Unsupervised Domain Adaptation (UDA) methods, such as those proposed in \cite{Luo_2019_ICCV,hoyer2022hrda,Zou_2018_ECCV}, have emerged as effective approaches for improving the performance of models on unlabelled target domain data. 
These methods rely on the availability of labelled source domain data during the adaptation. 
However, there are situations where access to the labelled source domain data is restricted due to privacy and security concerns. 
For instance, the labelled data may originate from sensitive consumer information, making it infeasible to release the data to third parties. 
In such cases, only the  models pretrained on the source domain data are accessible for adaptation, while the source domain data itself remains inaccessible.

The challenges presented by the restricted access to labelled source domain data have motivated us to propose a source-free domain adaptation approach specifically tailored for semantic segmentation tasks. 
Our work draws inspiration from previous research \cite{Yang2020eccv_labelrecon, Li2019bidirect} that explores the generation of additional data for domain adaptation. 
Similar to earlier studies \cite{jiang2020tsit,isola2017pix2pix,park2019SPADE},  we incorporate semantic information as a prior for generating realistic and diverse data.

In this paper, we introduce a novel source-free domain adaptation approach that specifically addresses the challenges associated with semantic segmentation tasks under this setting.
Since no ground-truth labels are available, the noisy predictions from the pretrained model reduce performance when used as labels during self-supervision.
Prior methods mitigate the detrimental effect of noisy labels by filtering \cite{Kundu2021genadapt} or loss rectification\cite{guo2022simt,zheng2021rect}.
However, loss rectification methods increase  computational overhead and potentially hinder training efficiency. 
Label filtering reduces the number of training examples available, and for imbalanced training datasets, also disproportionately affects the performance of minority classes compared to majority classes.

We address the limitation arising from the lack of ground-truth labels by deploying a framework that generates target-like images from the model predictions.
Instead of discarding uncertain predictions, we generate target-like images that are semantically consistent with the model predictions. 
Additionally, the target-like images are also stylistically consistent with the corresponding class in the model predictions.
This allows the model predictions to serve as the `ground truth' labels for the generated target-like images.

Figure \ref{figure:figure1_overview} illustrates the reasoning behind our approach.
The predictions from the pretrained segmentation model are semantically inconsistent with the ground truth and are not suitable to be used as labels for the original target domain images.
However, for our approach,  the semantic consistency between the generated target-like images and the model predictions enables the use of model predictions as labels for the target-like images.
At the same time, the `sidewalk' pixels that are incorrectly classified as `road' pixels have styles resembling that of `road' pixels.
This stylistic similarity between the target-like images and the original target domain images allows us to improve segmentation model performance by minimizing the stylistic differences between the target-like images and the original target domain images while training the segmentation model. 
The experimental results for the two experimental settings GTA5\cite{Richter_2016_ECCV}\textrightarrow Cityscapes \cite{Cordts2016Cityscapes} and Synthia  \cite{RosCVPR16}\textrightarrow Cityscapes \cite{Cordts2016Cityscapes} demonstrate the efficacy of our proposed solution.

We summarize our main contributions in this paper:
\begin{itemize}
    \item We introduce a simple yet novel image translation approach for the source-free domain adaptation 
    setting. 
    To our knowledge, our work is the first to generate target-like images from pixel-level pseudo labels under the challenging source-free domain adaptation setting.
    \item The target-like images are stylistically similar to the original target domain images while being semantically consistent with the noisy model predictions. 
    We then leverage the generated target-like images to improve adaptation performance.
    \item We demonstrate the effectiveness of our approach comparable performance with state-of-the-art work on two benchmark datasets. For example, our approach outperforms recent state-of-the-art work (Guo \etal \cite{guo2022simt} and Kundu \etal \cite{Kundu2021genadapt})  for GTA5\cite{Richter_2016_ECCV}\textrightarrow Cityscapes \cite{Cordts2016Cityscapes} by 0.6$\%$ and 1.8\% respectively. Our approach also demonstrates comparable results with Kundu \etal \cite{Kundu2021genadapt} for Synthia  \cite{RosCVPR16}\textrightarrow Cityscapes \cite{Cordts2016Cityscapes}.
    
\end{itemize}

\section{Related work}
\begin{figure*}[htp]
\includegraphics[width=1.0\textwidth]{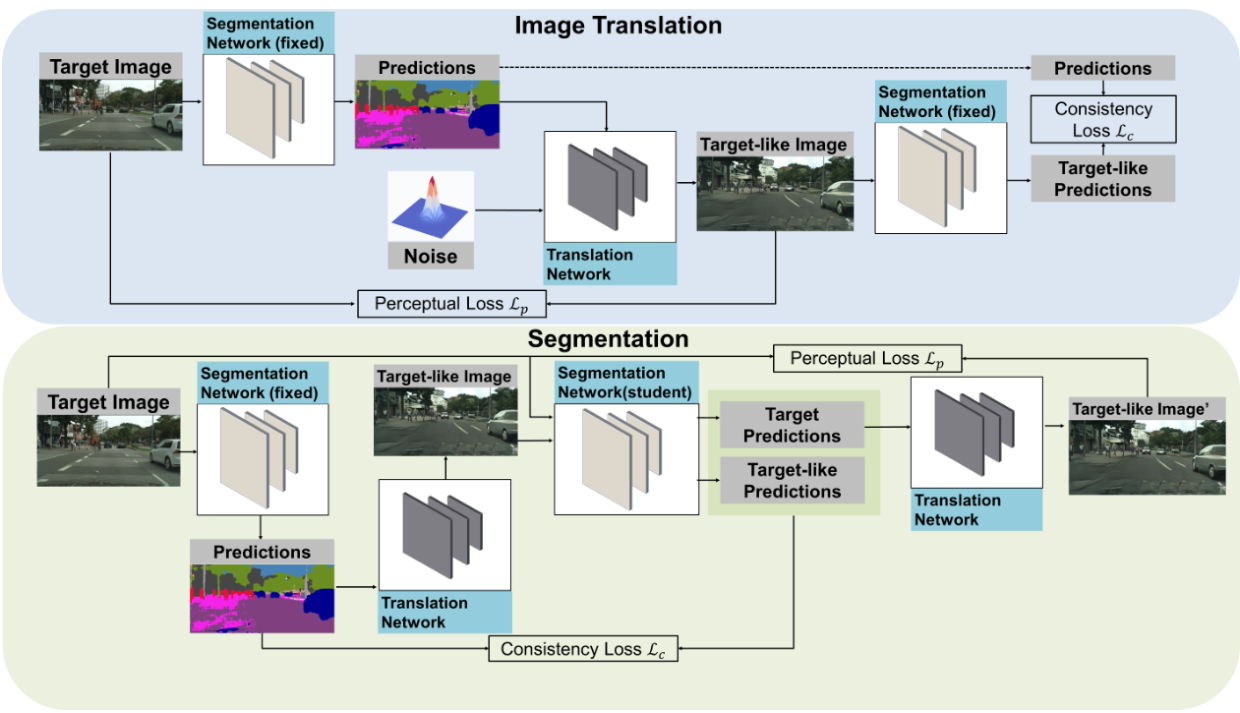}
 \vspace{-0.5cm}
\captionsetup{width=1.0\textwidth}
\caption{Illustrated workflow for ReGEN. We train the image translation network to generate target-like images that are semantically consistent with the one-hot encoded pseudo labels. For instance, the incorrectly classified pixels (\ie ``road''$\rightarrow$``sidewalk'' have been updated in the target-like image to resemble that of the sidewalk. This maximizes the consistency between the target-like images and the pseudo-labels, thus providing a reliable source of labels for supervision.
We train the segmentation network to minimize the perceptual loss between the target-like image'(the image generated with predictions from the student segmentation network) and the original target image, in addition to minimizing the consistency loss between the predictions from fixed teacher segmentation network and those of the student model.}
\label{figure:Workflow fig2}
\end{figure*}

\subsection{Source-free domain adaptation}
Source-free domain adaptation is a technique that aims to adapt a pretrained source model to an unlabelled target domain without using the source domain data during the adaptation process. 
Under this setting, resolving the challenges posed by noisy pseudo labels during self-training is essential.

A possible solution is to generate source-like/target-like data during adaptation.
While generating source-like data \cite{Liu_2021_SFDA, Hou2020imgtrans,Yang2020eccv_labelrecon} simplifies the source-free domain adaptation problem to an unsupervised domain adaptation problem, generating sufficiently diverse and representative source-like data still remains an open problem.
Liu \etal \cite{Liu_2021_SFDA} train a generator to output source-like images using input randomly drawn from a Gaussian distribution.
Hou \etal \cite{Hou2020imgtrans} first train a modified CycleGAN \cite{Zhu2017CycleGAN} on the source domain images to generate source domain images before adapting the model to generate source-like images from the target domain images. 
However, their method initially requires access to the source domain images to train the modified CycleGAN, which may not always be feasible for real-life applications.
Yang \etal \cite{Yang2020eccv_labelrecon} leveraged the labelled source domain images to generate source-like images via an image translation network.
For the generation of target-like images, Li \etal \cite{Li_2020_CVPR} explored the possibility of conditioning image generation with image-level labels for adapting image classification tasks.
Our approach differs from the above-mentioned work by generating target-like data without using any labelled data.

\subsection{Data generation via generative models} 
Generative methods, particularly Generative Adversarial Networks (GANs) \cite{Goodfellow2014GANS}, have demonstrated their effectiveness in a diverse range of computer vision applications, including super-resolution \cite{wang2018sftgan}, image-to-image translation \cite{Zhu2017CycleGAN}, and image denoising \cite{chen2018_denoisingGANs}. 
The success of GANs in these tasks has served as an inspiration for their use in addressing domain adaptation challenges.
Recently, conditional generative methods, as exemplified in \cite{Li_2020_CVPR, Yang2020eccv_labelrecon}, have shown the capability to synthesize target-like data based on a given prior, such as image-level or pixel-level labels. 
Li \etal \cite{Li_2020_CVPR} focused on generating target-like data for image classification using predefined image-level labels. 
However, applying this approach to semantic segmentation tasks becomes challenging due to the large number of pixels involved, making it infeasible to predefine pixel-level labels. 
In the work of Yang \etal \cite{Yang2020eccv_labelrecon}, they explored the generation of images based on pixel-level semantic information. 
Their approach constrained the translation network to generate images that are semantically consistent with the input by training the image translation network to generate source-like images from the predictions of the target domain images. 
However, their method \cite{Yang2020eccv_labelrecon} relies on labelled source domain data during training, which is unavailable in the source-free setting.
\section{Methods}
\subsection{Workflow}
\begin{algorithm*}  \caption{Pseudo code for ReGEN}
\begin{algorithmic}[1]
  \INPUT Pretrained teacher segmentation network $G_{fixed}$, Image translation model \textit{T}, Number of iterations to train translation model $Iter_{tr}$, Number of iterations to jointly train both the translation model and segmentation network $ Iter_{joint}$, Target domain images $\boldsymbol{X}_{tgt}$
  \OUTPUT Adapted segmentation network $G$ 
\FOR{$0$, ..., $Iter_{tr}$}
 \STATE Generate one-hot predictions $\boldsymbol{Y^{'}}\leftarrow G_{fixed}(\boldsymbol{X}_{tgt})$
 \STATE Generate target-like images $X^{'}_{tgt}\leftarrow T(\boldsymbol{Y^{'}})$
\STATE  Update $\theta_{T}$ via minimising $\mathcal{L}_{translation}\leftarrow \mathcal{L}_p(\boldsymbol{X}^{'}_{tgt},\boldsymbol{X}_{tgt}) + \mathcal{L}_c(\boldsymbol{X}^{'}_{tgt},\boldsymbol{Y^{'}})+ \mathcal{L}_f(\boldsymbol{X}^{'}_{tgt},\boldsymbol{X}_{tgt})+ \mathcal{L}_{f}(\boldsymbol{X}^{'}_{tgt},\boldsymbol{X}_{tgt}) + \mathcal{L}_{KLD} $
  \ENDFOR \FOR{$0$, ..., $Iter_{joint}$}
    \STATE Generate target-like images from pretrained teacher segmentation network $\boldsymbol{X}^{'}_{tgt}\leftarrow T(G_{fixed}(\boldsymbol{X}_{tgt}))$ 
    \STATE  Update $\theta_{T}$ via minimising $\mathcal{L}_{translation}\leftarrow \mathcal{L}_p(\boldsymbol{X}^{'}_{tgt},\boldsymbol{X}_{tgt}) + \mathcal{L}_c(\boldsymbol{X}^{'}_{tgt},\boldsymbol{Y^{'}})+ \mathcal{L}_f(\boldsymbol{X}^{'}_{tgt},\boldsymbol{X}_{tgt})+ \mathcal{L}_{f}(\boldsymbol{X}^{'}_{tgt},\boldsymbol{X}_{tgt}) + \mathcal{L}_{KLD}(\boldsymbol{X}^{'}_{tgt},\boldsymbol{X}_{tgt})$
   \STATE Filter one-hot predictions $\boldsymbol{Y}^{'}$ from $G_{fixed}(\boldsymbol{X}_{tgt})$ via class-wise confidence thresholding to get $\boldsymbol{Y}^"$.
   \STATE Generate target-like images from the student segmentation network $\boldsymbol{X}^{"}_{tgt}\leftarrow T(G(\boldsymbol{X}_{tgt}))$ 
   \STATE Update $\theta_{G}$ via minimising $\mathcal{L}_{seg}\leftarrow  \mathcal{L}_{p}(\boldsymbol{X}^{"}_{tgt},\boldsymbol{X}_{tgt}) + \mathcal{L}_{c}(\boldsymbol{X}_{tgt},\boldsymbol{Y}^")+ \mathcal{L}_{c}(\boldsymbol{X}^{'}_{tgt},\boldsymbol{Y}^{'}) + \mathcal{L}_{f}(\boldsymbol{X}^{"}_{tgt},\boldsymbol{X}_{tgt}) + \mathcal{L}_{KLD}(\boldsymbol{X}^{"}_{tgt},\boldsymbol{X}_{tgt})$ 

  \ENDFOR
 \end{algorithmic}
 \label{alg:workflow_alg}
\end{algorithm*}
Our workflow consists of two modules: an image translation network, denoted as \textit{T}, and a segmentation network, denoted as \textit{G}. 
The image translation network includes a generator, $T_{g}$, which generates target-like images from the segmentation network predictions, and a discriminator, $T_{D}$, which distinguishes between the original target domain images and the generated target-like images.
The overall workflow is visually depicted in Figure \ref{figure:Workflow fig2}. 

We first focus on the problem of generating suitable data for training.
As observed earlier by Kim \etal \cite{Kim_2020_CVPR}, simply translating the images by matching the colour distributions\cite{Hoffman_cycada2017} introduce blurring artifacts.
These artifacts degrade the semantic information present in the original images, making the translated images sub-optimal for training.
While it has been shown that pixel-level label information can be used to generate realistic images\cite{isola2017pix2pix, jiang2020tsit} and the feasibility of using pixel-wise label-driven image generation to address UDA problems \cite{Yang2020eccv_labelrecon} has already been demonstrated, we are the first, to the best of our knowledge, to generate target-like images from pseudo labels under the source free setting.

In order to generate target-like images, we enforce the constraint that the predictions from the target-like images are consistent with the predictions from the original target domain images (Figure  \ref{figure:Workflow fig2}).
This is done by minimizing the semantic consistency loss (Equation \ref{eqn:cross_ent_eqn}), which is simply the cross-entropy loss, while training the generator $T_{g}$.
Additionally, we also minimize the perceptual difference (Equation \ref{eqn:perceptual}) and the GAN feature matching loss (Equation \ref{eqn: feature matching loss}) between the target-like image and the target domain images.

We then adapt the segmentation network \textit{G}, which is pretrained on the labelled source domain data ($\boldsymbol{X}_{src}$,$\boldsymbol{Y}_{src}$) to the target domain using the unlabelled target domain data $\boldsymbol{X}_{tgt}$.
This is achieved by training the segmentation network \textit{G}, with generated target-like images and the original target domain images.

\subsection{Objective functions}
\textbf{Image translation} We first train the generator $T_{g}$ in the image translation network to generate target-like images from the one-hot predictions of the target domain images.

Following Jiang \etal\cite{jiang2020tsit}, we use the hinge-based adversarial loss \cite{lim2017geogan}, KL divergence loss \cite{park2019SPADE}, perceptual loss \cite{Johnson2016_ECCV_percept} (Equation \ref{eqn:perceptual}), semantic consistency loss (Equation \ref{eqn:cross_ent_eqn}) and GAN feature matching loss\cite{Wang_2018_CVPR_highres_imgsyn} (Equation \ref{eqn: feature matching loss}) to train the generator.
The discriminator is trained with hinge-based adversarial loss.

\textbf{Perceptual loss}
We apply perceptual loss \cite{Johnson2016_ECCV_percept} $\mathcal{L}_{p}$ to minimize the visual gap between the generated target-like images and the target domain images during the training of the image translation network in the first stage, followed by joint training of the segmentation network and the image translation model in the second stage (Algorithm \ref{alg:workflow_alg}).
Similar to Jiang \etal \cite{jiang2020tsit}, we minimize the L1 loss between the feature representations from the original target domain images and the target-like images. 
We extract the features from the following layers $\phi_{i}$ i.e. (\texttt{relu1\textunderscore1, relu2\textunderscore1, relu3\textunderscore1, relu4\textunderscore1, relu5\textunderscore1}) of the pretrained VGG19 network $\phi$, with the loss weights $w_{i}$ set at 1\slash32, 1\slash16, 1\slash8, 1\slash4, 1.

The perceptual loss $\mathcal{L}_{p}$ is given by the following equation:
\begin{equation}
\begin{aligned}
&\mathcal{L}_{p}(T_{g},G,\boldsymbol{X}_{tgt})= \\
&\sum_{i=1}^{5}w_{i}\lVert \phi_{i}(T_{g}(G(\boldsymbol{X}_{tgt})))-\phi_{i}(\boldsymbol{X}_{tgt})\rVert_{1},
\end{aligned}
\label{eqn:perceptual}
\end{equation}
In the first phase, we use the fixed segmentation network to generate predictions from the original target domain. 
We then generate the target-like images from those predictions.
Additionally, we use the perceptual loss to train the segmentation network in the joint training phase (Equation \ref{alg:workflow_alg}).
In the second phase, we use the predictions from the student segmentation network's instead of the fixed teacher segmentation network to generate target-like images $\boldsymbol{X}^{"}_{tgt}$.
Assuming a well-trained image translation network, any visual discrepancies between the target-like images and the original target images would be due to prediction errors from the student segmentation network.
This allows the perceptual loss to improve the performance of the segmentation network by leveraging the unlabelled target domain images.

\textbf{Semantic Consistency loss}
We determine the semantic consistency loss $\mathcal{L}_{c}$ between the input images $\boldsymbol{X}$ and the target domain images by computing the cross-entropy loss, as shown by the following:
\begin{equation}
\mathcal{L}_{c}(G,\boldsymbol{X},\boldsymbol{Y^{'}}) = \sum_{i=1}^{H \times W}\sum_{c=1}^{C} -Y^{'}_{ic}log(G(\boldsymbol{X})),
\label{eqn:cross_ent_eqn}
\end{equation}
where $G(\boldsymbol{X})$ refers to the predicted probability of class $c$ for the \textit{i}th pixel for the input image $\boldsymbol{X}$. $Y^{'}_{ic}$ is the predicted label by the fixed teacher segmentation network for class $c$ on the \textit{i}th pixel, where $Y^{'}_{ic}=1$ if the pixel belongs to the class \textit{c} and $Y^{'}_{ic}=0$ if otherwise.

Minimizing the cross entropy loss $\mathcal{L}_{c}(G_{fixed},\boldsymbol{X},\boldsymbol{Y^{'}})$ while freezing the weights of segmentation network $G$, will steer the image translation network to generate target-like images that are semantically consistent with the predictions $\boldsymbol{Y^{'}}$.

\textbf{GAN Feature Matching loss} The GAN feature matching loss\cite{Wang_2018_CVPR_highres_imgsyn} is similar to the perceptual loss\cite{Johnson2016_ECCV_percept}, though it compares the feature representations obtained from several discriminator $T_{D}$ layers. 
It is calculated as:
\begin{equation}
\begin{aligned}
&\mathcal{L}_f(\boldsymbol{X^{'}}_{tgt},\boldsymbol{X}_{tgt})= \\
&\mathbb E(\boldsymbol{X}_{tgt})\sum_{i}^{N}\lVert T_{D}^{(i)}(\boldsymbol{X}_{tgt})-T_{D}^{(i)}(\boldsymbol{X^{'}}_{tgt}) \rVert_{1} ,
\label{eqn: feature matching loss}
\end{aligned}
\end{equation}
where \textit{N} refers to the number of layers in the discriminator $T_{D}$ and $\boldsymbol{X^{'}}_{tgt}$ is the target-like image.
The overall loss function for training the image translation network is
\begin{equation}
\begin{aligned}
\mathcal{L}_{translation} =& \lambda_{p}\mathcal{L}_{p}(\boldsymbol{X}^{'}_{tgt},\boldsymbol{X}_{tgt}) + \lambda_{c}\mathcal{L}_{c}(\boldsymbol{X}_{tgt},\boldsymbol{Y}^{'})\\& +\lambda_{KLD}\mathcal{L}_{KLD}(\boldsymbol{X}^{'}_{tgt},\boldsymbol{X}_{tgt})\\& +\lambda_{f}\mathcal{L}_{f}(\boldsymbol{X}^{'}_{tgt},\boldsymbol{X}_{tgt}),
\label{eqn:combined_image_translation}
\end{aligned}
\end{equation},
where $\mathcal{L}_{KLD}$ refers to the KL divergence loss \cite{park2019SPADE} commonly used for generative tasks.
We filter the segmentation model predictions $\boldsymbol{Y}^{'}$ by selecting the top $33\%$ confident pixels per class to obtain the filtered pseudo labels $\boldsymbol{Y}^{"}$.
The hyperparameter weights used in our implementation are $\lambda_{c}=3.0$, $\lambda_{KLD}=0.05$, $\lambda_{f}=1.0$ and $\lambda_{p}=2.0$ for training the image translation network.

\textbf{Semantic Segmentation} The overall loss function for training the semantic segmentation network is
\begin{equation}
\begin{aligned}
\mathcal{L}_{seg} =&\lambda_{tgt}\mathcal{L}_c(\boldsymbol{X}_{tgt},\boldsymbol{Y^"}) + \lambda_{gen}\mathcal{L}_{c}(\boldsymbol{X}^{'}_{tgt},\boldsymbol{Y}^{'}) \\&
+\lambda_{pseg}\mathcal{L}_{p}(\boldsymbol{X}^{"}_{tgt},\boldsymbol{X}_{tgt})
+\lambda_{f}\mathcal{L}_{f}(\boldsymbol{X}^{"}_{tgt},\boldsymbol{X}_{tgt}) \\&
+\lambda_{KLD}\mathcal{L}_{KLD}(\boldsymbol{X}^{"}_{tgt},\boldsymbol{X}_{tgt}),
\label{eqn:combined_semantic}
\end{aligned}
\end{equation}
and the hyperparameter weights used in our implementation are $\lambda_{KLD}=0.05$,$\lambda_{tgt}=1.0$, $\lambda_{gen}=3.0$, $\lambda_{f}=1.0$ and $\lambda_{pseg}=10$ for training the segmentation network.
$\boldsymbol{X}^{"}_{tgt}$ refers to the target-like images generated from the student model predictions.

\section{Experiments}
In this section, we introduce the datasets used for training and evaluation of adaptation performance (Section \ref{subsection:datasets}), followed by the network architectures used for image translation and semantic segmentation (Section \ref{subsection:Network Architecture}).  
\subsection{Datasets}
\label{subsection:datasets}

Following prior source-free domain adaptation work \cite{Liu_2021_SFDA,Kundu2021genadapt}, we evaluate our proposed method with the following datasets.
\begin{itemize}
   \item \textbf{GTA5}\cite{Richter_2016_ECCV} is a synthetic semantic segmentation dataset with 24,966 densely annotated images with resolution $1914 \times 1052$ pixels, and has 19 categories that are compatible with the Cityscapes\cite{Cordts2016Cityscapes} dataset.
    \item \textbf{Synthia}\cite{RosCVPR16} refers to the SYNTHIA-RAND-CITYSCAPES subset from the publicly available database for semantic segmentation. 
    It has 9,400 densely annotated images with resolution $1280 \times 760$ pixels and has 16 categories that are compatible with the Cityscapes\cite{Cordts2016Cityscapes} dataset.
    \item \textbf{Cityscapes}\cite{Cordts2016Cityscapes} is a real-world driving dataset with densely annotated images of resolution 2048 $\times$ 1024 pixels. 
    We use the Cityscapes dataset as the target domain, following the default split of 2,975 unlabelled images: 500 images for training and evaluation of model performance respectively. 
\end{itemize}
\subsection{Network Architecture}
\label{subsection:Network Architecture}
Here, we introduce the network architecture involved in the image reconstruction and semantic segmentation tasks.
We implement our workflow with the Pytorch library \cite{paszke2015_pytorch}.

\textbf{Image translation}
For image translation, we use the simplified version of the two-stream image translation network $T$ \cite{jiang2020tsit}. 
Unlike Jiang \etal\cite{jiang2020tsit}'s approach where the generator contains a content-stream and style-stream module that allows for content and style inputs, we use a generator containing only the content-stream module to reduce the number of model parameters required.
We found no significant difference in performance by including the additional style input.
In our approach, the generator takes the one-hot encoded segmentation model predictions as input.
We use the multi-scale patch discriminator \cite{park2019SPADE}, based on the approach by Jiang \etal \cite{jiang2020tsit}.
We use the Adam optimizer\cite{DBLP:journals/corr/KingmaB14} with $\beta_{1}=0$, $\beta_{2}=0.9$.
The learning rate for the generator and the discriminator is $10^{-4}$ and $4\times10^{-4}$.
We first train the image translation network for up to 80 epochs with batch size=1. 
 
\begin{table*}[htp]
 \begin{center}
 \resizebox{\textwidth}{!}{%
 \begin{tabular}{p{2cm}| c|abababababababababab|c}
  \toprule
  \multicolumn{23}{c}{\textbf{GTA5 $\rightarrow$ Cityscapes}} \\
  \toprule & \rotatebox{90}{Year} & \rotatebox{90}{Arch.}
   & \rotatebox{90}{road} & \rotatebox{90}{side.} & \rotatebox{90}{buil.} & \rotatebox{90}{wall} & \rotatebox{90}{fence} & \rotatebox{90}{pole} & \rotatebox{90}{light} & \rotatebox{90}{sign} & \rotatebox{90}{vege.} & \rotatebox{90}{terr.} & \rotatebox{90}{sky} & \rotatebox{90}{pers.} & \rotatebox{90}{rider} & \rotatebox{90}{car} & \rotatebox{90}{truck}& \rotatebox{90}{bus} & \rotatebox{90}{train} & \rotatebox{90}{motor} & \rotatebox{90}{bike} &  \rotatebox{90}{\textbf{mIoU}} \\ 
   \midrule
 AUGCO \cite{2022arXiv_Zaugco} & 2022 & R & 90.3 & 41.2 & 81.8 & 26.5 & 21.4 & 34.5 & 40.4 & 33.3 & 83.6 & 34.6 & 79.7 & 61.4 & 19.3 & 84.7 & 30.3 & 39.5 & 7.3 & 27.6 & 34.6 & 45.9 \\

 SFDA\cite{Liu_2021_SFDA} & 2021 &R &   84.2 &  39.2 &  82.7 &  27.5 &  22.1 &  25.9 &  31.1 &  21.9 &  82.4 &  30.5 &  85.3 &  58.7 &  22.1 &  80.0 &  33.1 &  31.5 &  3.6 &  27.8 &  30.6 & 43.2\\
 SOMAN \cite{Kundu2021genadapt}& 2021& R & 91.3 & 52.8 & 85.7 & 38.5 & 31.3 & 35.2 & 37.3 & 35.3 & 85.8 & 46.1 & 88.6 & 60.4 & 32.4 & 86.1 & 54.9 & 51.1 & 5.8 & 41.8 & 50.7
& 53.2\\
 SimT \cite{guo2022simt} & 2022 &R & 92.3 & 55.8 & 86.3 & 34.4 & 31.7 & 37.8 & 39.9 & 41.4 & 87.1 & 47.8 & 88.5 & 64.7 & 36.3 & 87.3 & 41.7 & 55.2 & 0.0 & 47.4 & 57.6 & 54.4\\
SF \cite{paul2021unsupervised}  & 2022 & R & 89.2 & 37.3 & 82.4 & 29.0 & 23.5 & 31.8 & 34.6 & 28.7 & 84.8 & 45.5 & 80.2 & 62.6 & 32.6 & 86.1 & 45.6 & 43.8 & 0.0 & 34.6 & 54.4 & 48.8 \\
ReGEN (Our approach) & 2023 & R & 92.6 & 56.2 & 86.5 & 36.0 & 33.2 & 39.1 & 38.2 & 46.1 & 87.5 & 45.9 & 87.6 & 65.8  & 37.1 & 87.9  & 43.8 & 57.7 & 0.0 & 44.8 & 58.5
& \textbf{55.0}\\
\bottomrule
\end{tabular}}
\end{center}
\caption{Segmentation performance of Deeplab-v2 with ResNet-101 backbone trained on GTA5, adapted to unlabelled Cityscapes data.} 
\label{tab:gta-cityscapes-paper-proof}
\end{table*}
\begin{table*}[htp]
 \begin{center}
 \resizebox{\textwidth}{!}{%
 \begin{tabular}{p{2cm}|c|ababababababababa|cc}%
\toprule
  \multicolumn{20}{c}{\textbf{Synthia $\rightarrow$ Cityscapes}} \\
  \midrule &\rotatebox{90}{Year} & \rotatebox{90}{Arch.}
   & \rotatebox{90}{road} & \rotatebox{90}{side.} & \rotatebox{90}{buil.} & \rotatebox{90}{wall \#} & \rotatebox{90}{fence \#} & \rotatebox{90}{pole \#} & \rotatebox{90}{light} & \rotatebox{90}{sign} & \rotatebox{90}{vege.} & \rotatebox{90}{sky} & \rotatebox{90}{pers.} & \rotatebox{90}{rider} & \rotatebox{90}{car} & \rotatebox{90}{bus} & \rotatebox{90}{motor} & \rotatebox{90}{bike} &  \rotatebox{90}{\textbf{mIoU13}}& \rotatebox{90}{\textbf{mIoU16}} \\ 
   \midrule
AUGCO\cite{2022arXiv_Zaugco} & 2022 & R &74.8 & 32.1 & 79.2 & 5.0 & 0.1 & 29.4 & 3.0 & 11.1 & 78.7 & 83.1 & 57.5 & 26.4 & 74.3 & 20.5 & 12.1 & 39.3 & 39.2 & 45.5\\

SFDA \cite{Liu_2021_SFDA} & 2021& R &   81.9 &  44.9 &  81.7 & 4.0 &0.5 & 26.2 &  3.3 &  10.7 &  86.3 &  89.4 &  37.9 &13.4&  80.6 &  25.6 &  9.6   &  31.3& 39.2 & 45.9 \\
SOMAN* \cite{Kundu2021genadapt} & 2021& R &  89.7 & 50.2 & 81.8 & 14.0 & 2.9 & 35.9 &  27.9 &30.9 &84.0 &  88.8 &66.6 &34.6 & 84.0 &  52.7 &  46.1 &47.9&  60.4 & 52.5\\
SF\cite{paul2021unsupervised} & 2022& R &74.3 & 33.7 & 78.9 & 14.6 & 0.7 & 31.5 & 21.3 & 28.8 & 80.2 & 81.6 & 50.7 & 24.5 & 78.3 & 11.6 & 34.4 & 53.7 & 50.2 &43.7 
 \\

SimT \cite{guo2022simt} & 2022& R &  87.5 &37.0 &79.7 & 7.8 & 1.0 & 30.2 &9.5 &17.3 & 79.4 & 80.3 &53.4 &20.8 &82.0 &34.2 &18.5 &38.5& 49.1 & 42.3 \\

ReGEN (Our approach) & 2023& R  & 88.3 & 42.96 & 80.81 & 9.22 & 0.69 & 37.93 & 23.96 & 28.56 & 82.69 &
83.15 & 68.01 & 35.3 & 83.04 & 39.57 & 
42.5 & 54.89 & 58.0& 50.1 \\
\bottomrule
\end{tabular}}
\end{center}
\caption{Segmentation performance of Deeplab-v2 with ResNet-101 backbone trained on Synthia, adapted to unlabelled Cityscapes data. Note*: The reported score here is derived from the model checkpoint available on the project page maintained by Kundu \etal\cite{Kundu2021genadapt}. mIoU13 and mIoU16 are computed over 13 classes (excluding the classes marked with \#) and 16 classes respectively.}
\label{tab:synthia-cityscapes-paper-proof}
\end{table*}
\textbf{Semantic Segmentation}
We use the DeepLab-v2 \cite{chen2017_deeplab} segmentation network with ResNet-101\cite{He2015resnet} backbone for the segmentation model $G$. 
We use the pretrained weights for the segmentation models from Kundu \etal \cite{Kundu2021genadapt} and Guo \etal \cite{guo2022simt} for the Synthia and GTA5 datasets respectively.
Similar to Kundu \etal \cite{Kundu2021genadapt}, we freeze all the layers except for the layer preceding the classifiers in the segmentation model.
We use the SGD optimizer with momentum 0.9, an initial learning rate  $2.5\times10^{-4}$, a polynomial learning rate decay of power 0.8 and weight decay $5\times10^{-4}$. 

To ensure a fair comparison for the GTA5$\rightarrow$ Cityscapes setting, we used the same model checkpoint that Guo \etal\cite{guo2022simt} obtained after the initial warm-up stage. 
Similarly, to ensure a fair comparison for the Synthia$\rightarrow$Cityscapes setting, we use the same model checkpoint that Kundu \etal \cite{Kundu2021genadapt} obtained before the self-training step in their implementation.
We first perform 3 rounds of self-training on the target domain data following the approach by Kundu \etal \cite{Kundu2021genadapt} to warm up the pretrained segmentation model.
We then jointly train the segmentation model and the image translation network for a maximum of 50,000 iterations on a single NVIDIA A100 GPU card, with batch size=2.
During this phase, we filter the pseudo labels for the original target domain images using the class-wise confidence thresholding approach\cite{Zou_2018_ECCV,Kundu2021genadapt}. 
Similar to Kundu \etal \cite{Kundu2021genadapt}, we set the class-wise thresholds at $33\%$ of the most confident predictions at each iteration.
Pixels with prediction probabilities lower than the threshold are assigned to an `unlabelled' class and ignored during loss computation.

\begin{table}[ht]
\setlength{\tabcolsep}{12pt}
\centering
\begin{tabular}{c c c c c  }
\toprule
$\lambda_{c}$      & 3.0     & 3.0 & 6.0  & 9.0 \\
$\lambda_{p}$      & 2.0     & 4.0 & 2.0  & 2.0 \\
$\lambda_{f}$ & 1.0    & 2.0 & 1.0 & 1.0 \\
\midrule
Avg. mIoU          & 55.0    & 54.8
&  53.8  &   53.4  \\
\bottomrule
\end{tabular}
\captionsetup{width=\columnwidth}
\caption{Hyperparameter evaluation for the GTA5 $\rightarrow$ Cityscapes setting for training the image translation network $T$. The hyperparameters $\lambda_{c}$, $\lambda_{p}$ and $\lambda_{f}$ refer to the weights for the semantic consistency loss, perceptual loss, and GAN feature matching loss as shown in Equation \ref{eqn:combined_image_translation}. }
\label{tab:content, perceptual and ganfeat}
\end{table}
\begin{table}[ht]
\setlength{\tabcolsep}{12pt}
\centering
\begin{tabular}{c c c c  c }
\toprule
$\lambda_{tgt}$ & 1.0   & 1.0  & 3.0 & 3.0  \\
$\lambda_{pseg}$& 10.0  & 2.0  & 10.0 & 10.0 \\
$\lambda_{gen}$ & 3.0   & 3.0  & 3.0  &  0 \\
\midrule
Avg. mIoU       & 55.0  & 53.0 &  52.6  &  51.9     \\
\bottomrule
\end{tabular}
\captionsetup{width=\columnwidth}
\caption{Hyperparameter evaluation for the GTA5$\rightarrow$Cityscapes setting for training the image segmentation network $G$. The hyperparameters $\lambda_{tgt}$, $\lambda_{pseg}$ and $\lambda_{gen}$  refer to the loss weights for the semantic consistency loss for the target images $\mathcal{L}_{c}(\boldsymbol{X_{tgt}},\boldsymbol{Y}^{"})$, perceptual loss $\mathcal{L}_{p}$ and the semantic consistency loss for the target-like images $\mathcal{L}_{c}(\boldsymbol{X_{tgt}^{'}},\boldsymbol{Y}^{'})$ as shown in Equation \ref{eqn:combined_semantic}. }
\label{tab:content, perceptual and ganfeat semantic seg}
\end{table}

\begin{figure*}
\includegraphics[width=1.0\textwidth]{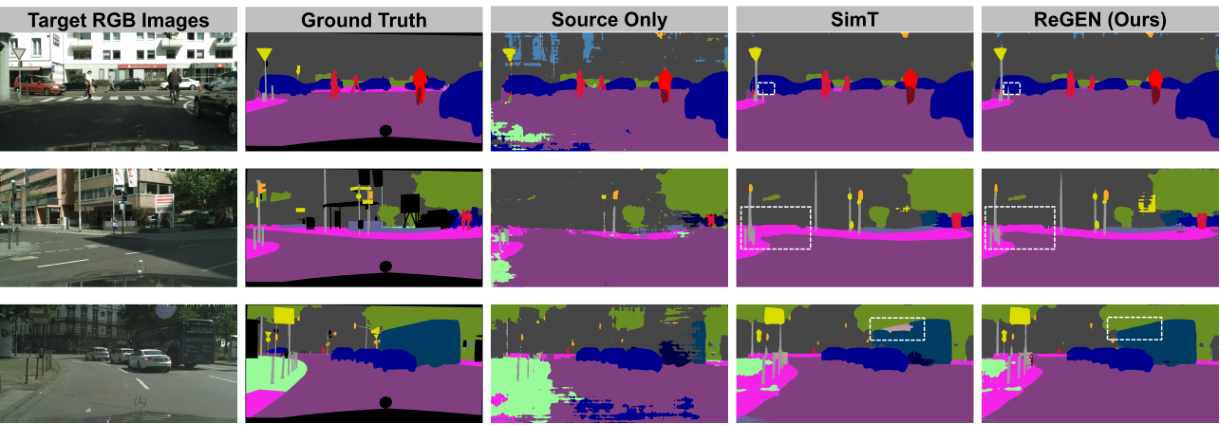}
 \vspace{-0.5cm}
\captionsetup{width=1.0\textwidth}
\caption{Qualitative results for the GTA5\textrightarrow Cityscapes setting. Our approach, ReGEN, demonstrates qualitatively better performance, being able to resolve small objects (top row: ``pole"), manage confusion cases (middle row: ``road"-``sidewalk") and avoid classification errors (bottom row: ``bus"-``fence").}
\label{figure:qualitative_results}
\end{figure*}

\section{Discussion}
In this section, we compare our work with prior art and also evaluate the hyperparameter weights used to train the image translation network and the semantic segmentation network.
\subsection{Comparison with prior work}
In Table \ref{tab:gta-cityscapes-paper-proof} and \ref{tab:synthia-cityscapes-paper-proof}, we compare our proposed approach, ReGEN, with the state-of-the-art work\cite{Kundu2021genadapt,guo2022simt} and also with representative prior work \cite{Liu_2021_SFDA,paul2021unsupervised,2022arXiv_Zaugco}.
Guo \etal \cite{guo2022simt} addresses the challenge of open-set semantic segmentation by learning a noise transition matrix that mitigates the effect of noise in the pseudo labels.
Kundu \etal \cite{Kundu2021genadapt} trains the segmentation network backbone and multiple classifier heads with differently augmented source domain data for each of the classifier heads to maximise model generalizability, followed by self-training with the unlabelled target domain data.
Liu \etal \cite{Liu_2021_SFDA} generate source-like data by leveraging the learned parameters of the pretrained segmentation network.
Paul \etal \cite{paul2021unsupervised} enforce consistency between the model output from several input pixel-level transformations of unlabelled target domain data.
Prabhu \etal \cite{2022arXiv_Zaugco} train the segmentation model to maximise consistency between the augmented target domain images, while also identifying reliable pseudo labels via class-conditioned confidence thresholding. 

Our proposed approach demonstrates comparable performance with state-of-the-art work for both experimental settings (Table \ref{tab:gta-cityscapes-paper-proof},\ref{tab:synthia-cityscapes-paper-proof}).
In particular, our approach surpasses all other methods for the GTA5$\rightarrow$Cityscapes setting and demonstrates comparable performance with state-of-the-art work for the Synthia$\rightarrow$Cityscapes setting.

We also present a qualitative comparison of our work in Figure \ref{figure:qualitative_results}.
Compared to the prior state-of-the-art work \cite{guo2022simt}, our approach demonstrates better performance resolving small objects (\eg pole, traffic sign) and distinguishing between the confusion classes (``road"-``sidewalk" and ``person"-``rider").

\subsection{Hyperparameter evaluation}
\textbf{Image Translation} Table \ref{tab:content, perceptual and ganfeat} shows the effect of the loss weights $\lambda_{c}$, $\lambda_{p}$
and $\lambda_f$ used during image translation on segmentation performance.
Here, maximizing the ability of the translation model to generate target-like images with high semantic consistency with the segmentation model predictions is required for effective adaptation of the segmentation model. 
The results suggest that balancing the  weights for semantic consistency and stylistic similarity is essential for generating high-quality data for training. 
Additionally, raising the weights for semantic consistency reduced adaptation performance (as seen in the rightmost columns of Table \ref{tab:content, perceptual and ganfeat}).
This was initially surprising because a higher semantic consistency between the pseudo labels and the generated target-like images would mean more reliable supervision.
However, we suggest that this increased semantic consistency could have been achieved at the cost of reduced stylistic similarity with the original target domain images. 
This might explain why the adaptation performance was reduced in both cases.

\textbf{Semantic Segmentation}
Table \ref{tab:content, perceptual and ganfeat semantic seg} shows the effect of the loss weights $\lambda_{gen}$, $\lambda_{tgt}$, $\lambda_{pseg}$
and $\lambda_{f}$ on the segmentation model performance.
Comparison of the performance between the two leftmost columns in Table \ref{tab:content, perceptual and ganfeat semantic seg} suggest that  perceptual loss can be effective as an additional means of supervision.
However, as expected, semantic consistency loss for target-like images is also essential for achieving good performance (rightmost column in Table \ref{tab:content, perceptual and ganfeat semantic seg}).
The results also show that increasing the weights for the semantic consistency loss (from 1.0 to 3.0) of the target domain images reduces performance (Table \ref{tab:content, perceptual and ganfeat semantic seg}) and we suggest that this might be caused by the noise in the pseudo labels.

\begin{table}[h]
\setlength{\tabcolsep}{20pt}
\centering
\begin{tabular}{c c c c c}
\toprule
$\mathcal{L}_{c}$ & $\mathcal{L}_{p}$ & $\mathcal{L}_{f}$  & mIoU \\
\hline
 \checkmark &  \checkmark & \checkmark & 55.0  \\
 \checkmark &  \checkmark &  & 53.0  \\
 \checkmark& &    
 \checkmark &  51.9 \\
 &\checkmark& \checkmark& 51.9  \\ 
 
\bottomrule
\end{tabular}
\captionsetup{width=\columnwidth}
\caption{Evaluation of the effect on performance by eliminating perceptual loss $\mathcal{L}_{p}$, semantic consistency loss for the generated target-like images $\mathcal{L}_{c}(\boldsymbol{X_{tgt}^{'}},\boldsymbol{Y}^{'})$  and GAN feature loss $\mathcal{L}_{f}$ during training of the segmentation network for the GTA5\textrightarrow Cityscapes setting.}
\label{tab:booktabs_ablation}
\end{table}

\subsection{Ablation study}
We explore the effect of perceptual loss, semantic consistency loss for the target-like images and GAN feature matching loss during training of the segmentation network (Table \ref{tab:booktabs_ablation}).
The results show that both perceptual loss and semantic consistency loss have more influence on model performance compared to GAN feature matching loss.

Additionally, we wanted to determine whether filtering the target-like images would have an effect on model performance since filtering is commonly used for most self-supervised methods.
The reasoning behind this was to determine w
Therefore, we  retain the top k\% confident pixels per class ($10\%, 25\%, 50\%,75\%,100\%$)  when we generate the pseudo labels for the target-like image (Table \ref{tab: filtering pseudo label}).
We observed an upward trend between the percentage of pixels retained in the generated target-like images, though the performance drops when $50\%$ of the target-like pixels are retained. 
This drop might be caused by a far greater number of incorrect labels than correct labels for the target-like images occurring between $25\%-50\%$ confidence across the classes. 
While the performance was best when all the pixels were retained during training ($100\%$), the lack of  any considerable difference in performance for the different filtering rates seems to suggest that even with extremely high filtering rates \ie $10\%$, model performance remains high.
\begin{table}
\centering
\begin{tabular}{c c c c c c }
\toprule
$\%$ of pixels retained & 10 & 25 & 50 & 75 & 100\\ 
\toprule
 mIoU16   &  49.0  & 49.1 & 48.9  & 49.2 & 50.1\\
\bottomrule
\end{tabular}
\captionsetup{width=\columnwidth}
\caption{Evaluation of the effect of retaining the top k$\%$ confidence predictions per class made by the pretrained teacher model to generate pseudo labels for training the segmentation model under the Synthia\textrightarrow Cityscapes setting.}
\label{tab: filtering pseudo label}
\end{table}

\subsection{Qualitative evaluation of target-like images}
\begin{figure}
\includegraphics[width=1.0\columnwidth,keepaspectratio]{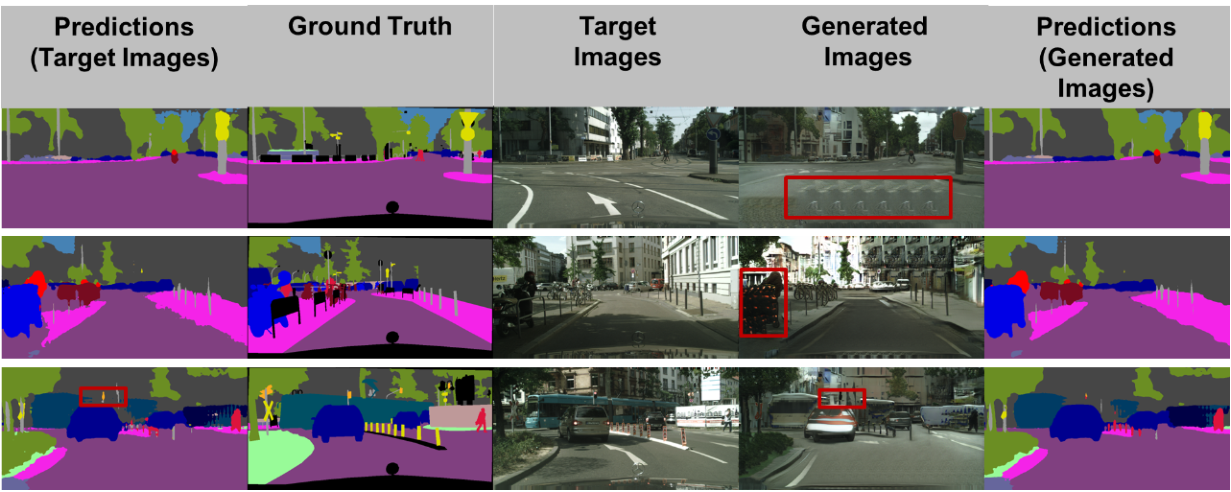}
 \vspace{-0.5cm}
\captionsetup{width=1.0\columnwidth}
\caption{Illustration of common error cases during image translation. Aliasing artifacts (top and middle rows) reduce the intra-class style diversity, which could reduce segmentation performance. Prediction errors (bottom row:`train'\textrightarrow `building') can cause the segmentation network to learn incorrect relationships between the pixels.  }
\label{figure:fig_failurecases}
\end{figure}

We observed that the generated target-like images show good semantic consistency with the input one-hot predictions (Figure \ref{figure:fig_failurecases}). 
As shown in the figure, the predictions from the generated images show good agreement with those of the original target images, despite some stylistic differences between the generated images and the original target domain images.
However, we also noticed some errors in the generated images.
Aliasing artifacts (characterized by unwanted repetitive patterns in the generated images) reduce  intra-class diversity in the target-like images.
These artifacts may affect performance on the original target domain images as the segmentation model may overfit to the generated instances.
\section{Conclusion}
We introduce a source-free domain adaptation workflow that generates target-like data with reliable pixel-level labels.
Our approach generates target-like data that has high semantic consistency while also possessing high stylistic similarity to the target domain images. 
For future work, we intend to further extend our workflow to address additional domain adaptation settings.

{\small
\bibliographystyle{ieee_fullname}
\bibliography{egbib}
}

\end{document}